# An Ensemble Model for Face Liveness Detection

Avinash Patel
Senior CV Scientist
AI Labs, Subex Ltd.
avinash.patel@subex.com

Mrinal Haloi
Principal CV Scientist
AI Labs, Subex Ltd.
mrinal.haloi@subex.com

Asif Salim
Research Scientist
AI Labs, Subex Ltd.
asif.salim@subex.com

Shashank Shekhar
Head
AI Labs, Subex Ltd.
shashank.shekhar@subex.com

*Abstract*— **In this paper, we present a passive method to detect face presentation attack a.k.a face liveness detection using an ensemble deep learning technique. Face liveness detection is one of the key steps involved in user identity verification of customers during the online onboarding/transaction processes. During identity verification, an unauthenticated user tries to bypass the verification system by several means, for example, they can capture a user photo from social media and do an imposter attack using printouts of users face or using a digital photo from a mobile device and even create a more sophisticated attack like video replay attack. We have tried to understand the different methods of attack and created an in-house large-scale dataset covering all the kinds of attacks to train a robust deep learning model. We propose an ensemble method where multiple features of the face and background regions are learned to predict whether the user is bonafide or an attacker.**

*Keywords—face liveness detection, face presentation attack, face recognition, convolutional neural network, ensemble learning*

## I. INTRODUCTION

The identity verification of customers in online applications has become ubiquitous with the boom of internet services. For this purpose, it is important to avoid any fraudsters misusing the identity of others for any illegal advantages. Hence as a major security measure, face recognition is an important and unavoidable step in providing the online services that carry financial risks. As part of this, the users are most often asked to turn on the camera of the devices and the system does the verification. During this process, it is important to identify whether the person is a real user that matches the credentials stored in the system or an imposter trying to fool the system.

There are several ways an imposter can bypass the security measures and those processes are collectively named as face presentation attacks. There are multiple ways an attack can happen. The common choices are (1) video replay attack in which the video of the person is played, (2) face mask attack in which the imposter wears a mask of another person's face being printed on it, (3) printed photo attack in which the attacker places a 2-D printed photo in front of the device, etc. To rectify these, it is crucial to detect the liveness of the facial features that has been given to the device and hence the face liveness detection is an important step for identity verification of online services.

There are solutions for the liveness detection that dates back at least three decades. Before the deep learning era, the solutions were proposed based on hand crafted features like the ones based on texture analysis extracted from the images. However, these methods were good at detecting attacks that are weak like a printed photo that clearly distinguishes the features of bonafide and attacking cases. The complexity of the attacks has also drastically improved by using video replay, photo realistic masks etc. With the advent of CNN, several methods were proposed as a solution to these sophisticated attacks.

In this paper, we describe a CNN based ensemble strategy for face liveness detection. The objective of our work is to solve the liveness problem through a series of simple steps or simple learning problems rather than having complex solutions. The state-of-the-art solutions in this aspect involve complex procedures as part of their learning and they may be mostly suited for tackling only a single type of attack. Through our ensemble learning strategy we could cover the cross attacks and we are able to propose a simple and robust model that is suitable for industrial use-cases.

The rest of the paper is organized into the following sections. Section II briefly discusses about the related works. In section III, the applications of the face liveliness detection are discussed. The proposed method is discussed in Section IV. The related experiments and discussions are provided in section V and the concluding remarks are given in section VI.

## II. RELATED WORKS

The related works can be divided into the ones before deep learning era and the ones after that. The earlier works were focused on extracting hand-crafted features for the task and they are mostly based on learning a binary categorization – whether the face is live or fake. The hand-crafted features were based on the image properties such as color, blurriness, texture, specular reflection etc. The examples of such works are [1], [2], [3], [4].

The advancements in computer vision with the advent of convolutional neural networks (CNN) also brought the development of new techniques to detect face presentation attack. Yang et.al [6] solved face presentation attack problem as a binary classification problem to classify either as real user or attacker by taking the features learned by a CNN. Gan et.al [5] analyzed the spatiotemporal features of continuous video frames using 3D-CNN to detect face anti-spoofing and they were able to detect video level attack compared to the solutions at that time which mainly focused on photo level attack. Shao et.al [7] used CNN to learn the facial textures to differentiate between mask attacks and real users. They extracted features in a channel wise manner and analysed optical flow for the purpose. Li et.al [8] proposed a solution based on 3D-CNN using which spatiotemporal features are analyzed. A patch-based CNN approach was proposed by Li et.al [9] in which image is partitioned into individual blocks and each block are then processed by a separate CNN.

Another class of proposed solution includes processing an extra information content apart from the image frames. For example, Liu et,al [10] proposed a CNN-RNN model to extract an auxiliary supervision along with the conventional binary classification cues. The auxiliary supervision is to estimate the face depth feature and to estimate rPPG signals. Tian et.al [11] uses a polarization imaging sensor along with the conventional image capture mechanism and uses this extra information to detect the attacks. Pan et.al [12] proposed a model that detects eye blinks of the user to detect liveness with additional components to detect the temporal changes around the eyes.

A method that can be categorized into the family of ensemble learning is proposed by Wen et.al [13]. In this method, the customized features such as specular reflection, blurriness, chromatic moment, and color diversity are used to create vector space embeddings. These embeddings are then fed into an ensemble learner consisting of multiple SVM classifiers.

By analyzing state-of-the-arts we can see that most of the methods have complex steps involved in the learning strategy, they are time consuming and most of them are developed to tackle one or two specific types of attack. The complex learning creates additional overheads and do not generalize well to the cross domains. For industrial scale applications, we need to build techniques that are fast and user friendly without compromising the performance. It is in this context that we propose our method in which there are only CNN based classifiers. We learn a set of global and local features and they are utilized to train individual CNN models in an ensemble learning framework. These features are the ones that are created around the different parts of the face. In the experiments, we found that these features were good enough to provide robust performance.

III. APPLICATIONS OF LIVENESS DETECTION

In this section, we discuss the applications of the liveness detection and its importance for the secure running of online applications.

All financial institutions especially banking/stock market/insurance are benefitting hugely from digitalization by cutting down time and resources, where these organizations are trying to bring every process over online platform starting from product marketing, customer acquisition, and client onboarding. Previously in the customer onboarding/registration process, user had to visit the vendors branches physically, understand the process, submit and verify their identity which was a bottleneck in the process. To ease the customer onboarding process, many organizations have come up with e-KYC (electronic – know your customer) solutions where a new customer can verify himself via remote channels without visiting the physical offices. An e-KYC solution typically involves submission of user identity data and verification where users need to authenticate themselves by uploading their selfie images. The e-KYC solution verifies if the user's face is matching with the photo in the identity documents.

In such e-KYC process, sometimes a fake user may try to present himself as someone else to use the vendors application on behalf of some other user. This is where our face liveness detection solution comes to rescue preventing such face presentation attacks and stop the fake users from bypassing the identity verification in e-KYC applications.

IV. PROPOSED METHOD

In this section, we discuss the data collection procedures, and our methodology in detail.

*A. Data Collection*

We analyzed the dataset requirements and collected data focusing on user's face for both real as well imposter attacks coming from different sources and types, for example, camera type (mobile/web), user ethnicity, image acquisition environment, illumination etc. We have also explored several publicly available datasets like CelebA-spoof [14] to explore and understand the various kinds of face presentation attacks and based on that we have created our own dataset with high quality images involving more robust real-world variations.

We asked members across our team from globe to record a selfie video while doing different activities. It helped us to collect selfie videos of users from a variety of conditions. We identified all the possible sources of face presentation attacks and classified them into 4 categories, namely, 1. printed photo, 2. digital photo, 3. video replay attack, and 4. card mask. The data corresponding to these attacks is created from the real user's data.

For each type of face presentation attack, we created dataset using different cameras and under different environments. The figure 1 demonstrates a sample of a real user's face and each type of face presentation attacks.

*B. The ensemble model.*

Our proposed ensemble model contains a bunch of individual CNN models that learn a corresponding specific feature related to the task. The features are chosen in such a way that they will be capable of distinguishing between different types of face presentation attacks.

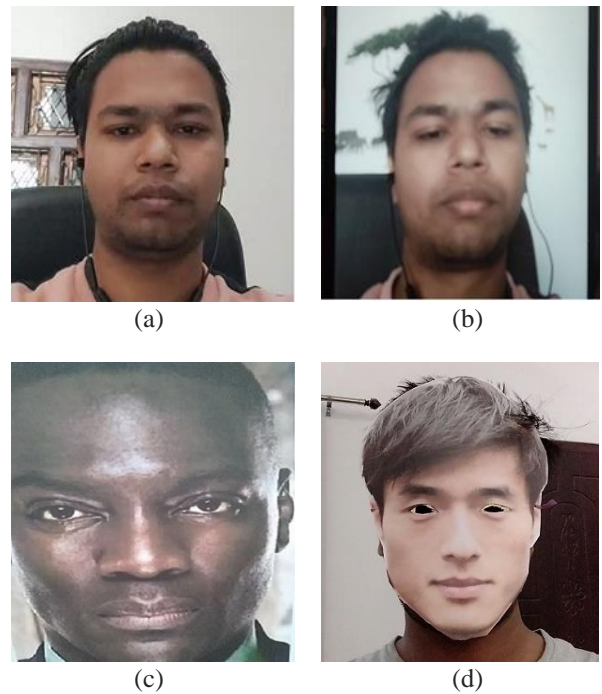

Fig 1: Examples of face presentation attacks. (a) real or bonafide user, (b) digital photo/ video replay attack, (c) printed photo attack, (d) card mask attack.

The objective is to solve the problem without complex learning strategies and rather to solve it through a collection of light CNN models. The steps involved in the learning procedure of an individual model in the ensemble is summarized in figure 2. A feature is extracted from the input image, and it is fed into a CNN model. The feature output from the CNN model is then fed into a classifier in which the image is classified either as a bonafide one or an attacker one.

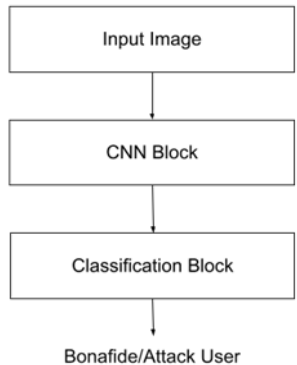

Fig 2: Data flow diagram of the individual CNN model in the ensemble

### C. Testing the liveness application in real-time users

In this section, we summarize the process of verification of the user to assign her as either bonafide or an attacker. To test the user's face liveness, we collect user's selfie video and select the frame with best quality. For selecting the best quality frame containing user's face we considered multiple image properties.

The frame selected is given as input to the individual model in the ensemble. The decision from the individual models is then aggregated to make the final decision.

## V. EXPERIMENTS

In this section, we describe the details of the experiments.

### A. Data augmentation

Generally, the training dataset which we use for machine learning model building might not be representing all the real-world scenarios, so we apply augmentation to the training dataset like random horizontal flips and random crops, and this helps in reducing the gap between real-world dataset and training dataset feature distributions. We have applied various augmentation techniques to cover different real-world scenarios in our training datasets and our empirical results show that deep learning model is more robust on the real-world dataset in testing step.

### B. Loss function

For our model training, the objective being the binary classification task, we have used binary cross entropy as loss function.

### C. Optimizer

While training our deep learning models we experimented with several optimizers to change the attributes of neural network to reduce the training losses. We find that for our dataset, Adam optimizer can learn the optimal model parameters quickly as compared to any other optimizers and avoids high variance in the model parameters during training.

### D. Regularization

We used dropout in the fully connected dense layers to the model from overfitting on the training dataset.

### E. Feature map Visualization using Grad-CAM

Deep learning model has facilitated unprecedented accuracy in various computer vision task. But one of the biggest problems with deep learning model is its interpretability/explain-ability which makes is hard to explain why model fails/succeeds. Recently researchers have tried several approaches to develop certain algorithms to visually debug the deep learning models and to find where the model is focusing in the image to make its decision. Grad-CAM is one of them. We used Grad-CAM to visually inspect where our deep learning classification model is looking into to identify the bonafide/attack users.

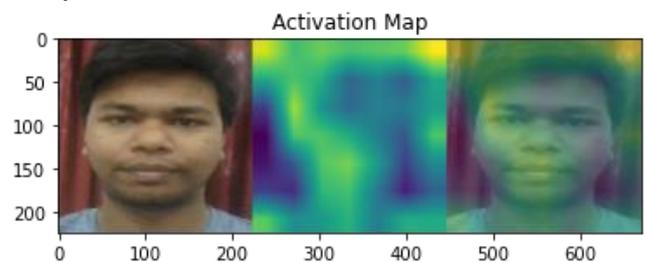

Fig 3(a): Grad-CAM based class activation map visualization of bonafide user – Model has identified relevant features from both background and face region.

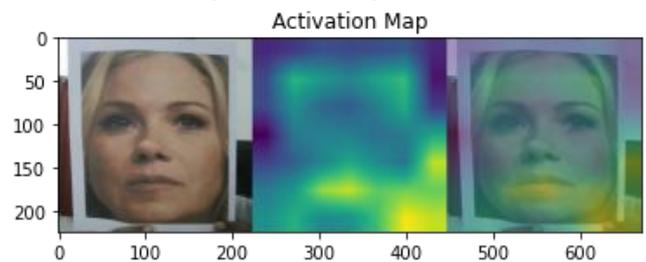

Fig 3(b): Printed Photo attack – Model has considered the photo edge regions to identify it as attack user.

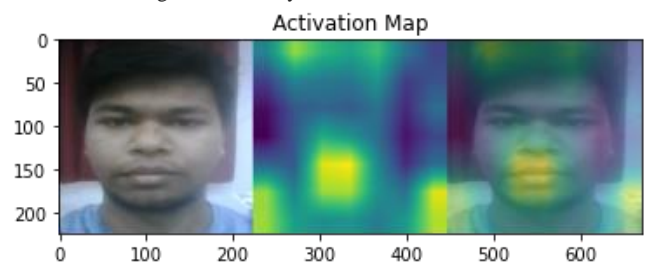

Fig 3(c): Replay Attack – Model has focused in the blurred regions to identify it as attack user.

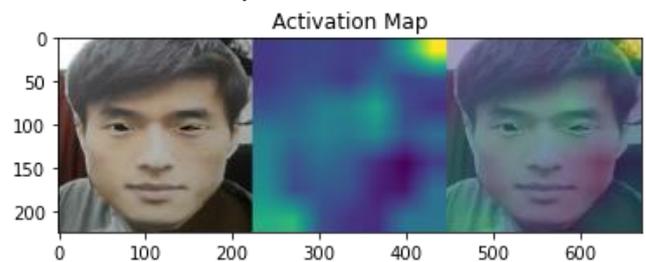

Fig 3(d): Card mask attack – Model has identified anomaly in the hair region to identify the image as card mask attack.

Fig 3 shows the examples of Grad-CAM based class activation map of input images for real and attack users. From the figures, we can see that the model has learned to process the right areas that differentiates between the bonafide and attacker cases.

| Real user cases | |
|---|---|
| Scenario | No: of cases |
| Indoor user in different lighting conditions and background | 4 |
| Indoor user in natural light and different background | 4 |
| Outdoor user with different background | 4 |
| Indoor and outdoor user with spectacles | 2 |
| **Attacking user cases** | |
| Scenario | No:of cases |
| Indoor user with paper mask | 2 |
| Outdoor user with paper mask | 2 |
| User with paper mask in artificial light | 2 |
| User with mobile photo in artificial light | 1 |
| User with mobile photo in natural light | 1 |
| User with plastic mask | 1 |
| User with shades on | 1 |
| Replay attack in artificial light | 1 |
| Replay attack in indoors | 1 |
| User with safety mask in indoor | 1 |
| User with safety mask in outdoor | 1 |
| User with safety mask with nose area cut in indoor | 1 |
| User with safety mask with nose area cut in outdoor | 1 |
| 2-D printed photo attack with African face | 1 |
| 2-D printed photo attack with American face | 1 |
| 2-D printed photo attack with Indian face | 1 |
| 2-D printed photo attack with Chinese face | 1 |
| Paper mask attack with African face | 1 |
| Paper mask attack with American face | 1 |
| Paper mask attack with Indian face | 1 |
| Paper mask attack with Chinese face | 1 |

Table 1: Testing cases of the ensemble model

*E. Evaluation Metrics*

To measure the performance liveness application in real world we have used three different metrics to measure the performance for both real and attack users. They are explained as follows.
1. APCER - Attack presentation classification error rate: The metric APCER is defined as FP/(TN+FP) where FP is false positive, and TN is true negative.
2. BPCER - Bonafide presentation classification error rate: This metric is defined as FN/(TP+FN) where FN is false negative, and TP is true positive.
3. ACER - Average classification error rate ie. Misclassification rate of classifier: This metric is defined as the average of APCER and BPCER.

*F. Results*

In this section, we discuss the results obtained with the models we built.

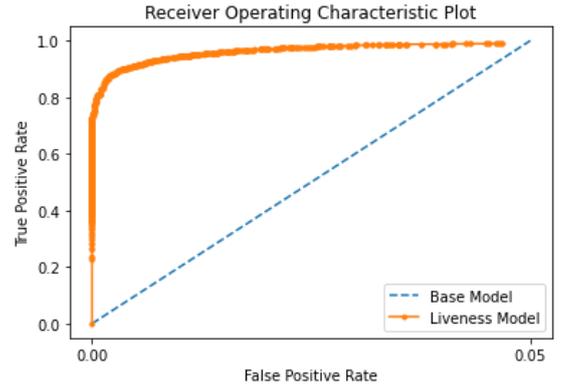

Fig 3: ROC plot of the proposed model

The ROC plot and precision v/s recall plots of the proposed model is given in figure 3 and 4 respectively. From the plots we can see that our model has robust performance in predicting both the bonafide and attacking cases.

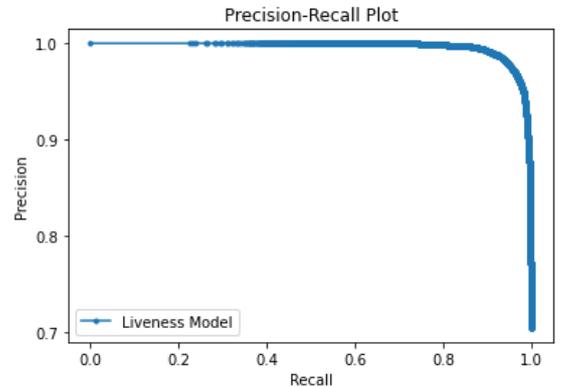

Fig 4: Precision- recall plot of the proposed model

The proposed ensemble model is subjected to rigorous testing conditions. We chose 6 subjects as both real users and attackers. We chose 14 different testing case for the real users and 24 cases for the attacking case for different scenarios as explained in table 1. So, a total of 228 testing cases (6 x (14+24)) are applied.

Based on the 228 testing cases, the results for the ensemble models are summarized in table 2.

| Metric | Score |
|---|---|
| APCER | 0 % |
| BPCER | 5.355 % |
| ACER | 2.67 % |

*Table 2: Performance metric of the ensemble model*

The results are superior compared with the state-of-the-art models as we subjected the proposed ensemble models for rigorous testing cases.

## VI. CONCLUSIONS

In this paper, we have presented an ensemble model on the face liveness detection. We have explained about our dataset preparation strategies, model training, and different experiments to achieve the best performing model. We have also explained about the user selfie image acquisition on different devices. From our empirical results, we show that our model can identify the attack users with negligible friction to the bonafide users. Hence our model is suitable for deploying in industry applications where the customer experiences are highly valued without affecting the performance.